# Playing Flappy Bird via Asynchronous Advantage Actor Critic Algorithm

Elit Cenk Alp[1]    Mehmet Serdar Güzel[2*]

[1,2]*Department of Computer Engineering*
Ankara, Turkey
Corresponding author email: mguzel@ankara.edu.tr

*Abstract*— **Flappy Bird, which has a very high popularity, has been trained in many algorithms. Some of these studies were trained from raw pixel values of game and some from specific attributes. In this study, the model was trained with raw game images, which had not been seen before. The trained model has learned as reinforcement when to make which decision. As an input to the model, the reward or penalty at the end of each step was returned and the training was completed. Flappy Bird game was trained with the Reinforcement Learning algorithm Deep Q-Network and Asynchronous Advantage Actor Critic (A3C) algorithms.**

## 1. INTRODUCTION

Flappy Bird made a very fast entry into the market. It was the most downloaded mobile game at the beginning of 2014. But within a very short time the market has withdrawn. Flappy Bird game is a single player game. There is only one action that jump. The game is about deciding when a bird should jump. The bird ends as soon as it strikes down, up, or another object. While the bird is flying, the pipes come across and the user tries to jump the bird between the two pipes. Although the game is played with a single button, it is very difficult. A simple frame from Flappy Bird is shown in Figure 1.Reinforcement training can be done with Q-learning algorithm [1]. In the Q-learning algorithm, an award is given for each decision. Rewards are given for things that need to be done and penalties are given for things that should not be done. Each decision is made in a table. In the table, the status of the environment (game), the decision taken, the award received and whether the environment continues (game-over or not) are kept. Subsequent decisions are made through this stored data.

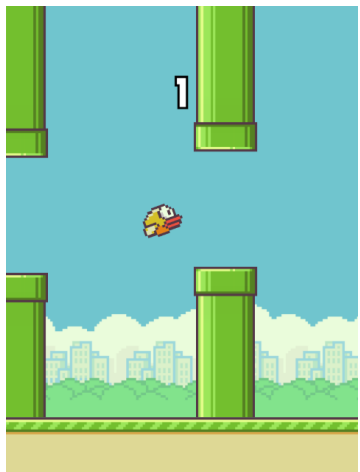

Fig 1. The raw image of the Game

In the Q-learning, the number of states increases and the number of actions to be taken increases. This takes a great deal of time to create the table in multivariable problems. There is a need to reduce the size of the table. The more advanced version of the Q-learning algorithm, the Deep Q-Network algorithm (DQN), has emerged to solve these problems. Mnih et al. [2] [3] made a huge impact in 2013. They developed an application that plays old arcade games with the DQN algorithm. They made an agent playing all the games in the arcade. There is no need to develop agent for each game separately. In the Q table, the some frames in the game were combined and the prizes were kept. Updating the rewards is similar to the Q-learning algorithm. The only difference is that the features in the game are taken from the images. Very successful results were obtained with this algorithm. Many games have been solved with DQN algorithm. Deep Networks have the ability to decide and remember. But training takes a long time. This training takes a long time and causes various problems. This is why the need to further develop the DQN algorithm. Prizes and penalties are predetermined in the DQN algorithm. In the Actor Critic algorithm, rewards can vary in each decision after they are pre-determined. Because when you play a game, the decisions you make are not always the same prize or punishment. Depending on the situation. As in reality, the reward in the Actor Critic algorithm can vary from case to case. Decision making in the DQN algorithm depends on the deep network model, as well as on the reward network. There is a model that makes decision making process and another model that produces a payment for the decisions of this model. Mnih et al. made an improvement that further accelerated the Actor Critic [4] algorithm. With the algorithm they have developed, asynchronous training can be done through agents faster. The name is called the Asynchronous Advantage Actor Critic algorithm. It is shown as A3C (Actor Critic algorithm which is non-asynchronous is also abbreviated as A2C). Previously, DQN algorithms take very long periods, while A3C can take much shorter periods. In this study, Flappy Bird game was trained using DQN algorithm followed by A3C [4] algorithm. The A3C algorithm is several times faster than the DQN algorithm. One week with DQN, GPU training, and the A3C algorithm takes one day with CPU. In the next section of the article, previous studies will be discussed. In the next section, the methods used will be explained. After the methods used, the experiments performed in the experiments section will be explained in detail. In the final section, the results section, the results will be discussed.

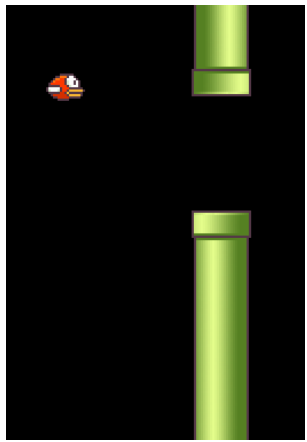

Fig 2. The pre-processed image of Flappy Image

2. **RELATED WORKS**

In 2013, Mnih et al. [2] developed an algorithm called DQN. In this study, the agent is trained from the images that have never been seen before. The algorithm that is tested on Atari games has produced results far above human results. Appiah and Vare [5] trained Flappy Bird with DQN in their study. They have found much better human results than before. In his study, Chen [6] trained Flappy Bird using various variations of the Q-Learning algorithm and the DQN algorithm. Ebeling-Rump and Hervieux-Moore [7] trained the game by applying Q-Learning variations.In 2016, Rosset et al. [8] trained with multi-agent Reinforcement

Learning. In Signh blog [9], made an A3C-learned agent that playing Flappy Bird. A comparison is not easy in training with reinforcement. Because the same environment may not be the same for every article. OpenAI has developed an environment to make environments better and make comparisons better. They called this environment Gym [10]. There are various environments. The applications have written can be easily tested on Gym environment. As well different AI based studies to different problems can be seen in [11,12,13,14].

## 3. METHOD

The reinforcement learning environment is PyGame. Flappy Bird game was run with PyGame. The environment in which the software is developed is Python. Deep Learning models are written in Keras, which is currently working on Tensorflow, and their training is done through these libraries.

### A. *Q Learning*

It determines the reward for performing a specific action when and in certain situations. It is a technique that evaluates which action should be performed based on an action-reward function.

Before starting the learning phase, Q - function (action-reward table) is started with a value determined by the programmer. Usually, 0 value is selected. Then, one action at a time is selected and a reward is returned to this action and the next situation is passed. Q – Function is updated after this action selection. The basis of the algorithm is to update the old reward value according to the new reward. Simple Q-learning algorithm shown in Figure 3. The new award is used to update the relevant record in the previously registered Q - function. After the previous Q - value is multiplied by a certain percentage, the new reward is added. This is shown in Equation 1.

$$Q(s_t, a_t)^{new} = (1 - \alpha) x Q(s_t, a_t)^{old} + \alpha x ( r_t + \gamma x \max Q(s_{t+1}, a_t)) \quad (1)$$

Where, at each time 't' the agent selects an action "$a_t$", observes a rewards "$r_t$" and encounters a newer state "$s_{t+1}$". In addition, while $\alpha$ refers learning rate, $\gamma$ illustrates discount factor.

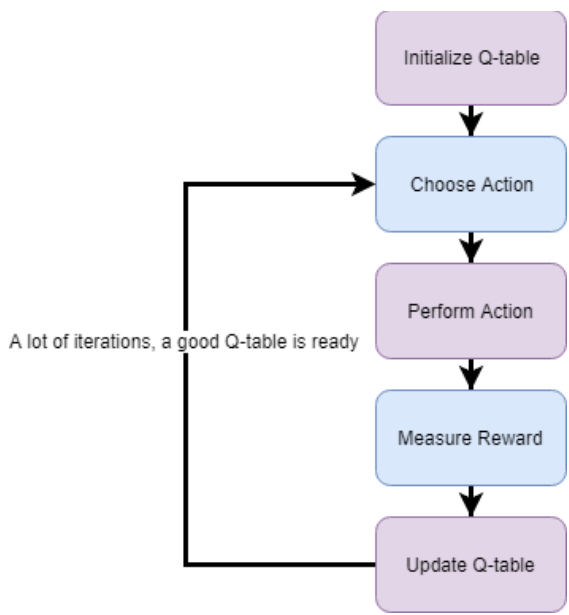

Fig 3. The Flow Chart of Q Learning Algorithm

## B. Deep Q-Network (DQN)

The Q - function is shown by a neural network. It is extremely useful for learning attributes from data with many features. The images corresponding to the last four images from game are taken as input and the Q - value is returned. To update the value of Q - function or to select the highest value of Q - function, the next state is estimated from the network. Since the Q - function is initially empty, it will perform a random search, and the training will continue until a successful strategy is found. The first successful method is not the most successful. Better methods can be found. This can be said to be greedy due to random search. The rate of selecting random values from the creation of the Q - function should be reduced over time, keeping the rate high. In a sense, it is correct to follow more random methods when creating history.Technique of experience replay [3] was used to stabilize the Q-value value over time. Technique of experience replay consists of storing a last learned value. Simple convolutional agent showed in Figure 4. The deep network structure using DQN is as follows. The first hidden layer convolves 32 filters of 8x8 with stride 4 and applies ReLU activation function. The second hidden layer convolves 64 filters of 4x4 with stride 2 and applies ReLU activation function. The third hidden layer convolves 32 filters of 3x3 with stride 1 and applies ReLU activation function. Last hidden layer consists 512 fully connected ReLU nodes. Final output layer has two nodes which is same number as Flappy Bird game action (Flap, Not Flap). 0th index node of output layer connected to Not Flap action. Values of the output layer represent the Q function given the input state for each action. At each step, the DQN model performs the best (highest) Q value. In training phase, replay memory with max size is 50000. Epsilon number is used in training with action performed random or not. First epsilon value is 0. 1. Epsilon value decreases as training progresses. Batch size is selected 32. In Q function updating, gamma value is 0.99. At each 1000 training step, we save the trained model. Adam was used as an optimizer. 1e-4 was used as learning rate.

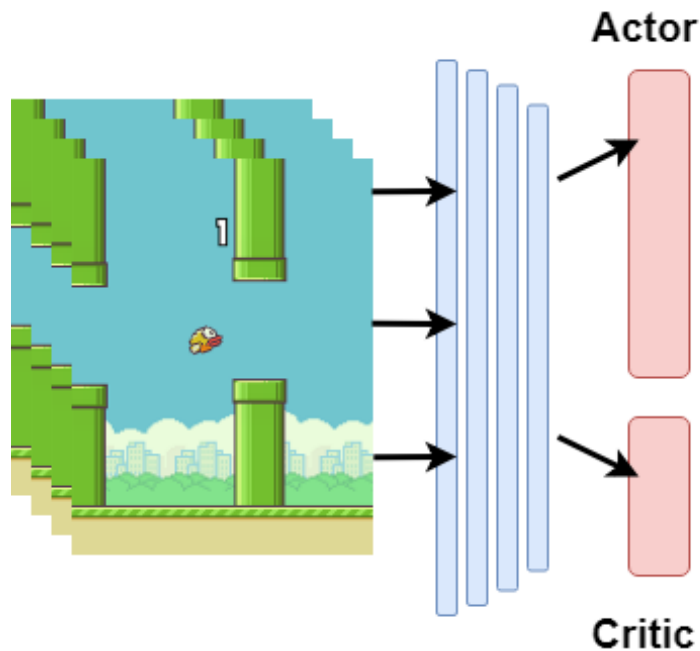

Fig 4. Actor Critic Model

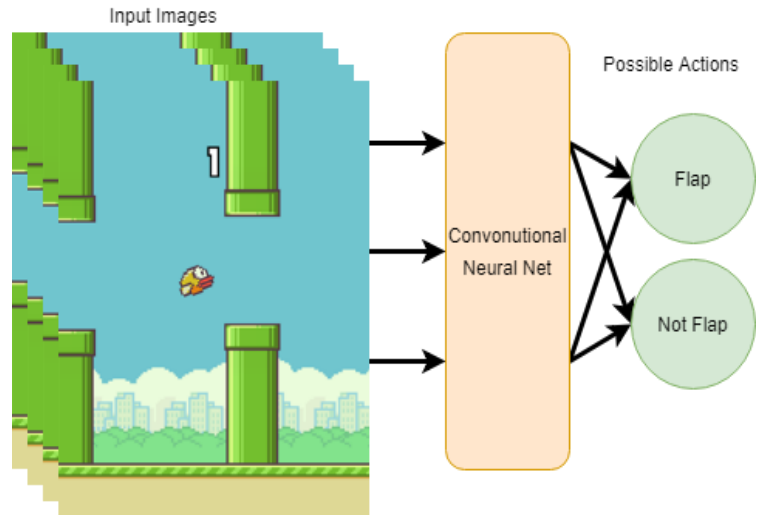

Fig 5. Convolutional agent

*C. Asynchronous Advantage Actor Critic (A3C)*

When we play a game, a fixed reward system is not always enough for us. For example, in a flappy game, if we give the same prize every time we pass a pipe, we get the same score from our first pipe pass and our pipe pass in the more difficult sections. Instead of a fixed reward system, setting up a system in which the prize can change will give better results. This was taken into account by an algorithm called Actor Critic in DeepMind's article [4]. In Deep Network model, instead of a single action output, a second output is added and the prize will be returned from that output. In this way, the award is given to a logic of reason and more accurate results are given. Actor Critic model shown in Figure 5. In addition to the Actor Critic model, DeepMind proposed an asynchronous structure. It has been shown that if multiple trainings are performed with no single agent training, faster results will be given. Flappy Bird, trained with GPU in one week with DQN algorithm, could be trained in six hour with A3C algorithm. Asynchronous Advantage Actor Critic model shown in Figure 5.

The advantages of the A3C over Deep Q-learning are:

- While the DQN learning time is too long (~ 1 week on the GPU), the A3C takes 1 day to run on a CPU. Flappy Bird training lasted six hours.

- DQN uses experience repetition to achieve convergence. This process uses a lot of memory. A3C uses threads to reduce this memory usage.

- A3C updates are made only from data generated from the current policy. DQN is a non-policy learning algorithm that can be stored in repetition of experience from data generated by an old policy, even if a better policy was discovered later. [9]

- The deep network used in the A3C model is as follows. Input of the neural network consist 84x84x4 images. The first hidden layer convolves 16 filters of 8x8 stride 4 and applies ReLU activation function. The second hidden layer convolves 32 filters of 4x4 stride 2 and applies ReLU activation function. Last hidden layer is 256 ReLU units. The output layer contains two type of output.

- **Policy Output:** This output value determines the actions in the model. This output value uses the sigmoid function. Since there are two actions in Flappy Bird, the output value of 1 is sufficient. The output value is between 0 and 1. The closer to 1, the greater the chance of flap making.

- **Critic Output:** This output value in the artificial neural network determines the reward corresponding to the input in this model. The output value is used in training by giving feedback to the network. The network is retrained if there is less than the prize value estimated by the critic. If the critic gives less chance of its occurrence, the situation is less likely to occur. This is how the A3C will award higher prizes in the future

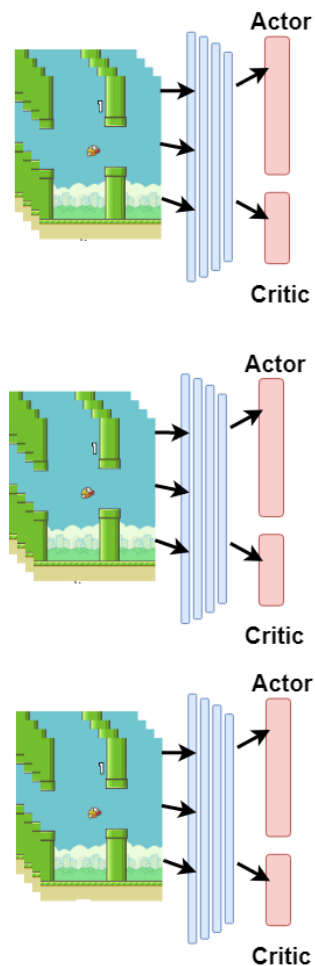

Fig 6. Asynchoronous Advantage Actor Critic model

### 4. EXPERIMENTS

Tensorflow was selected in the backend of the Keras library. Pygame was used to develop Flappy Bird game on Python. After the game, this environment was used for education. In the game, each pipe is given a point of 1 point and the bird is hit as a result of the death of -1 point penalty. Various modifications and experiments were performed in Flappy Bird. First, the sounds are removed. Secondly, the bird was prevented from being of different colours. Third, a fixed background colour was selected. Fourthly, in the education of the pictures coming from the game, not the whole game, but the reduced version of this picture was used. After the image is reduced, it is flipped in grey. The last image is highlighted according to their density.Two different experiments were performed in this study. DQN was used in the first experiment. This experiment lasted 1,500,000 iterations. As a result of so many iterations he has played as well as a human being. The average result of the experiment using DQN is shown in Figure 7. A3C, which is another experiment, has achieved much faster results. DQN training lasted quite slowly. Working with threads in A3C has made the result much faster. The experiment carried out on A3C lasted 9500 parts. Played 5 times in each thread. It was run with a total of 16 threads. In other words, a total of 16x5 = 80 games were played during one episode.

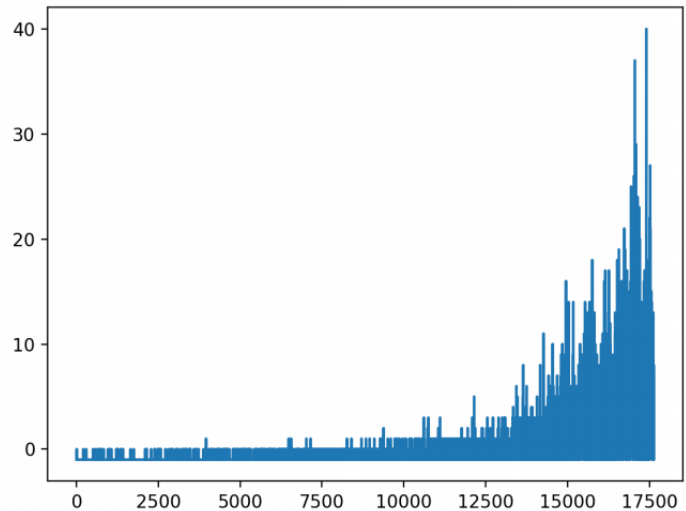

Fig 7. DQN avarage score.

## 5. CONCLUSIONS

Flappy Bird, DQN and A3C were trained in this paper. Very little information (images in the game) was used during the training. Experiments with the A3C resulted in much faster training. In addition to being fast, the A3C has been shown to deliver much better results. The reason why A3C gives better and faster results is that there is a reward system that can change in any situation. In the DQN algorithm, the prize is fixed from the environment. However, in A3C, it is determined by a neural network. In future studies, A3C method will be tested by adding HMM. Finally, the award system in the DQN method is also thought to be trained.